\documentclass[10pt,twocolumn,letterpaper]{article}

\usepackage{wacv}
\usepackage{times}
\usepackage{epsfig}
\usepackage{graphicx}
\usepackage{amsmath}
\usepackage{amssymb}

\usepackage{comment}
\usepackage{color}

\usepackage{microtype}
\usepackage{epsfig}
\usepackage{lipsum}

\usepackage{booktabs}
\usepackage{multirow}
\usepackage[normalem]{ulem}
\usepackage{appendix}
\usepackage{subcaption}

\def\wacvPaperID{1079} 

\wacvfinalcopy 
\ifwacvfinal
\usepackage[breaklinks=true,bookmarks=false]{hyperref}
\else
\usepackage[pagebackref=true,breaklinks=true,colorlinks,bookmarks=false]{hyperref}
\fi

\begin{document}

 \title{Deep Template-based Object Instance Detection}
 
\author{Jean-Philippe Mercier, Mathieu Garon, Philippe Giguère and Jean-François Lalonde\\
Université Laval, Québec, Canada\\
}

\maketitle
\thispagestyle{empty}

\begin{abstract}
Much of the focus in the object detection literature has been on the problem of identifying the bounding box of a particular class of object in an image. Yet, in contexts such as robotics and augmented reality, it is often necessary to find a \emph{specific} object instance---a unique toy or a custom industrial part for example---rather than a generic object class. 
Here, applications can require a rapid shift from one object instance to another, thus requiring fast turnaround which affords little-to-no training time. 
What is more, gathering a dataset and training a model for every new object instance to be detected can be an expensive and time-consuming process. In this context, we propose a generic 2D object \emph{instance} detection approach that uses example viewpoints of the target object at test time to retrieve its 2D location in RGB images, without requiring any additional training (i.e. fine-tuning) step. 
To this end, we present an end-to-end architecture that extracts global and local information of the object from its viewpoints. The global information is used to tune early filters in the backbone while local viewpoints are correlated with the input image.
Our method offers an improvement of almost 30 mAP over the previous template matching methods on the challenging Occluded Linemod~\cite{brachmann2014learning} dataset (overall mAP of 50.7). 
Our experiments also show that our single generic model (not trained on any of the test objects) yields detection results that are on par with approaches that are trained specifically on the target objects.

\end{abstract}

\section{Introduction}

\begin{figure}[t]
\begin{center}
\includegraphics[width=0.95\linewidth]{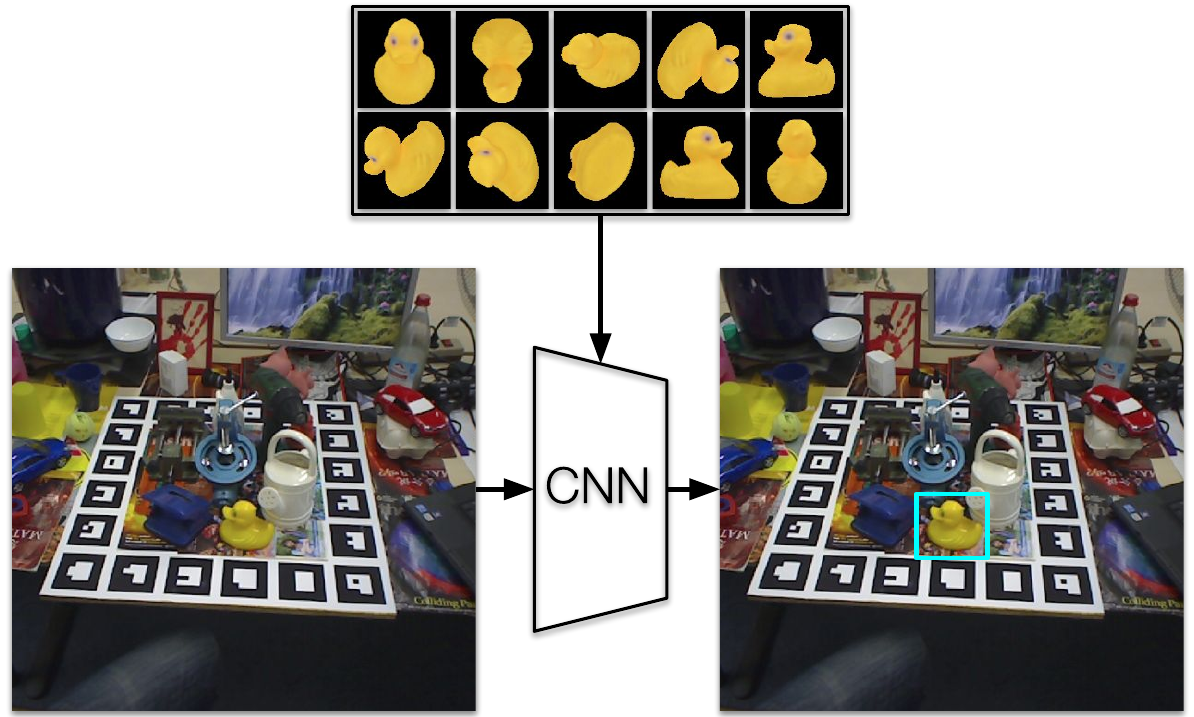}
\end{center}
\vspace{-0.17in}
\caption{Overview of the proposed method. At test time, our network predicts the 2D location (in an RGB image) of a target object (unseen during training) represented by templates acquired from various viewpoints.}
\label{fig:teaser}
\end{figure}

\vspace{-0.5em}
Object detection is one of the key problems in computer vision. While there has been significant effort and progress in detecting generic object classes (e.g. detect all the phones in an image), comparatively little attention has been devoted to detect specific object instances (e.g. detect \emph{this particular} phone model). Recent approaches on this topic~\cite{rad2017bb8,xiang2018posecnn,zakharov2019dpod,kehl2017ssd} have achieved very good performance in detecting object instances, even under challenging occlusions. By relying on textured 3D models as a way to specify the object instances to be detected, these methods propose to train detectors tailored for these objects. Because they know the objects to be detected at training time, these approaches essentially \emph{overfit} to the objects themselves: they become specialized at detecting them (and only them). 

While this is a promising and active research direction, requiring knowledge of the objects to be detected at training time might not always be practical. For instance, if a new object needs to be detected, the entire training process must be started over. This implies first generating a full training dataset and then optimizing the network. Also, using a single network per object can be a severe constraint in embedded applications where memory is a limited resource.

In this work, we explore the case of training a \emph{generic} 2D instance detector, where the specific object instance to be detected is only known at test time. The object to be found is represented by a set of images of that object captured from different viewpoints (fig.~\ref{fig:teaser}). In order to simplify the data capture setup and to facilitate comparisons to previous work on standard datasets, in this work we employ 3D models of the test objects and render different viewpoints. If a 3D model is not accessible, it would be possible to instead capture a few viewpoints of the object on a plain background.

This paper is akin to a line of work which has received somewhat less attention recently, that of template matching. These techniques scan the image over a dense set of sub-windows and compare each of them with a template representing the object. A canonical example is Linemod~\cite{hinterstoisser2011multimodal}, which detects a 3D object by treating several views of the object as templates, and by efficiently searching for matches over the image. While very efficient, traditional template matching techniques can be quite brittle, especially under occlusion, and yield large amounts of false positives. 

In this paper, we revive this line of work and propose a novel instance detection method. Using a philosophy sharing resemblance to meta-learning~\cite{vinyals2016matching}, our method uses a large-scale 3D object dataset and a rendering pipeline to learn a versatile template representation. 
At test time, our approach takes as input multiple viewpoints of any object and detects it from a single RGB image immediately, without any additional training (fig.~\ref{fig:teaser}).

Our main contribution is the design of a novel deep learning architecture which can localize instances of a target object from a set of input templates.
Instead of matching pixel intensities directly such as other template matching methods, our network is trained to localize an instance from a joint embedding space.
Our approach is trained exclusively on synthetic data and takes a single RGB image as input. In addition, we introduce a series of extensions to the architecture which improve the detection performance such as tunable filters to adapt the feature extraction process to the object instance in the early layers of a pretrained backbone. We quantify the contribution of each extension through a detailed ablation study. Finally, we present extensive experiments that demonstrate that our method can successfully detect object instances that were not seen during training. In particular, we report performances that significantly outperform the state of the art on the popular Occluded Linemod~\cite{brachmann2014learning} dataset. Notably, we attain a mAP of 50.71\%, which is almost 30\% better than LINE-2D~\cite{hinterstoisser2011multimodal} and on par with methods that overfit on the object instance during training.

\section{Related work}

\begin{figure*}[t]
\centering
\includegraphics[width=.85\linewidth]{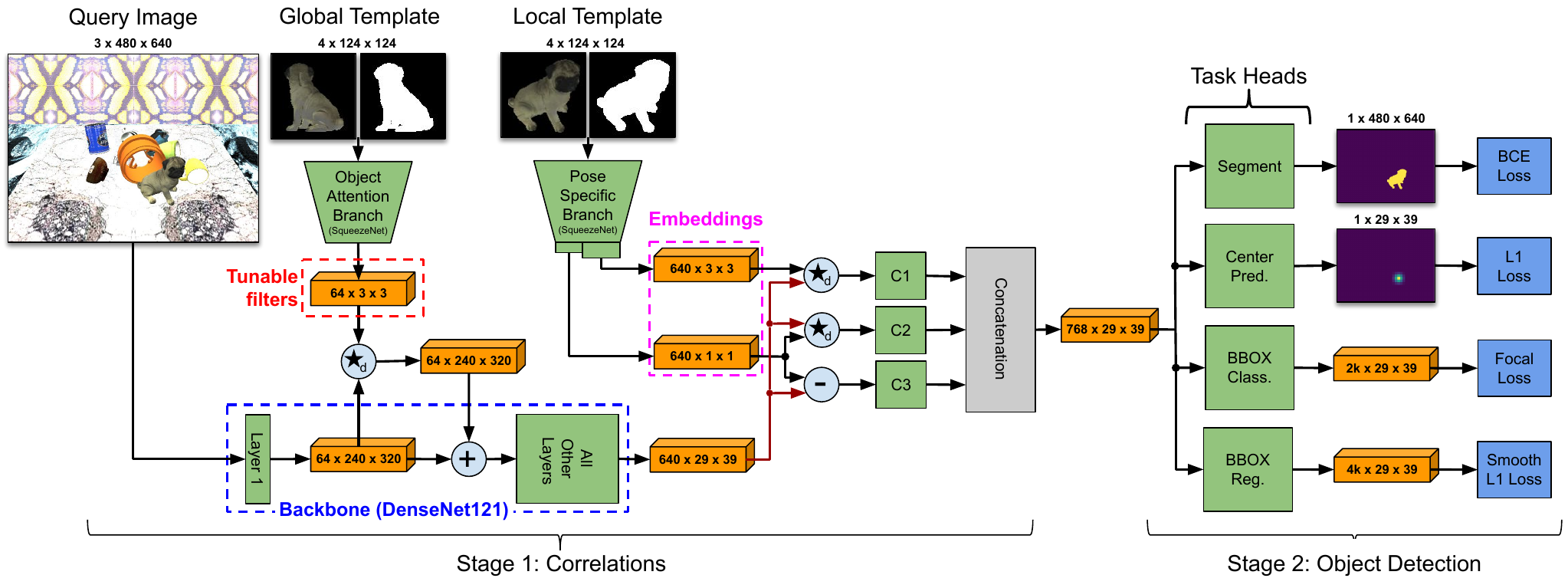}
\vspace{-0.1in}
\caption{Our proposed architecture. In stage 1, the network learns to localize an object solely from a set of templates. Object-specific features are learned by the ``object attention'' and ``pose-specific'' branches, and are subsequently correlated/subtracted with the generic features of the backbone network. In stage 2, the network leverages the learned representation to perform different tasks: binary segmentation, center and bounding box prediction. At test time, a single ``global'' template is randomly selected, while several ``local'' templates are combined.}
\label{fig:architecture}
\end{figure*}
\vspace{-0.5em}
Our work is most related to two areas: object instance detection in RGB images, and 2D tracking in RGB images. 

\vspace{-1em}
\paragraph{Object instance detection.} 
Our work focuses on retrieving the 2D bounding box of a \emph{particular object instance}.
This is in contrast with well-known methods such as Faster-RCNN~\cite{ren2015faster} and SSD~\cite{liu2016ssd} or with methods that bear more resemblance to our approach such as CoAE~\cite{hsieh2019one}, which all provide 2D locations of \emph{object classes}.
Detecting a specific object is challenging due to the large variety of objects that can be found in the wild. 
Descriptor-based and template-based methods are useful in such context. Generic features such as gradient histograms~\cite{hinterstoisser2011multimodal} and color histograms~\cite{tjaden2017real} can be computed and then retrieved from an object codebook.

Recent progress in deep learning enabled the community to develop approaches that automatically learn features from the 3D model of an object using neural network~\cite{rad2017bb8,xiang2018posecnn,zakharov2019dpod,kehl2017ssd} or random forest~\cite{brachmann2016uncertainty} classifiers. While these methods perform exceptionally well on known benchmarks~\cite{hodan2018bop}, they share the important limitation that training these deep neural networks requires a huge amount of labeled data tailored to the object instances to be detected. Consequently, gathering the training dataset for specific objects is both costly and time-consuming. Despite this, efforts have been made to capture such real datasets~\cite{brachmann2014learning,hinterstoisser2012model,hodan2017t,rennie2016dataset,doumanoglou2016recovering,tejani2014latent} and to combine them together~\cite{hodan2018bop}. A side effect is that it confines most deep learning methods to the very limited set of objects present in these datasets, as the weights of a network are specifically tuned to detect only a single~\cite{kehl2017ssd} or a few instances~\cite{kehl2017ssd,rad2017bb8}. The difficulty of gathering a real dataset can be partially alleviated using simple rendering techniques~\cite{hinterstoisser2018pre,kehl2017ssd,rad2017bb8} combined with data augmentation such as random backgrounds and domain randomization~\cite{tobin2017domain,zakharov2019deceptionnet,tremblay2018training},
but still suffers from a domain gap with real images. Recently, Hodan et al.~\cite{hodan2019photorealistic} demonstrated that the domain gap can be minimized with physics-based rendering. 
Despite this progress, all of the above methods share the same limitation, in that they all require significant time (and compute power) to train a network on a new object. This implies a slow turn-around time, where a practitioner must wait hours before a new object can be detected. 

To circumvent these limitations, we propose a novel generic network architecture that is trained to detect a target object that is unavailable at training time. Our method is trained on a large set of objects and can generalize to new, different objects at test time. Our architecture bears resemblance to TDID~\cite{ammirato2018target} that uses a template to detect a particular instance of an object. We show in our experiments that our method performs significantly better than \cite{ammirato2018target} on objects not seen during training.

\vspace{-1em}
\paragraph{Tracking in 2D images. }
Our work shares architectural similarities with 2D image-based tracking, for which approaches use a template of the object as input, typically identified as a bounding box in the first frame of the video. In contrast, we focus on single frame detection. Thus, we employ known viewpoints of the object acquired offline.
Many of these tracking approaches propose to use an in-network cross-correlation operation (sometimes denoted as $\star_d$) between a template and an image in feature space~\cite{wang2019fast,dave2019learning,li2019target}. Additionally, recent 6-DOF trackers achieve generic instance tracking using simple viewpoint renders from a 3D model~\cite{garon2018framework,li2018deepim,manhardt2018deep}. These methods are limited by the requirement of a previous temporal state in order to infer the current position. Our method takes inspiration from these two lines of work by first using the in-network cross-correlation and second, our experiments show that using renders is sufficient to \emph{locate} a specific object instance from a single RGB image.

\section{Network architecture}
\label{sec:architecture}
\def\pt{p_\textrm{t}}
\def\at{\alpha_\textrm{t}}
\def\FL{\textrm{FL}}

\vspace{-0.5em}
We first introduce an overview of our proposed network architecture, depicted in fig.~\ref{fig:architecture}. Then, we discuss the two main stages of our architecture: 1) correlation and 2) object detection. The correlation stage borrows from classical template matching methods, where the template of an object is compared to the query image in a sliding-window fashion. The second stage is inspired from the recent literature in class-based object detection.

\subsection{Architecture overview}
\vspace{-0.5em}
We design an architecture that receives knowledge of the object as input, computes the template correlation as a first stage, and regresses bounding boxes around the object from the correlation results in a second stage. As shown in fig.~\ref{fig:architecture}, the network takes as input the RGB query image and two types of templates: 1) a \emph{global} template used as an object attention mechanism to specialize early features in the backbone network; and 2) a \emph{local} template that helps extract viewpoint-related features. Each template is an RGB image representing the rendered 3D object from a given viewpoint on a black background, concatenated with its binary mask to form four-channel images. The templates are obtained with a fast OpenGL render of the object with diffuse reflectance, ambient occlusion and lit by a combination of one overhead directional light and constant ambient lighting.

\subsection{Correlation stage}
\label{sec:correlation_stage}
\vspace{-0.5em}
The query image is first processed by a conventional backbone to extract a latent feature representation. The global template is fed to an ``Object Attention Branch'' (OAB), which injects a set of tunable filters early into this backbone network such that the features get specialized to the particular object instance. On the other hand, the local template is consumed by the ``Pose-Specific Branch'' (PSB) to compute an embedding of the object. The resulting features are then correlated with the backbone features using simple cross-correlation operations. Note that at test time, the backbone (85\% of total computing) is processed only once per instance, while the second stage is computed for each template.

\vspace{-1em}
\paragraph{Backbone network.}
The role of the backbone network is to extract meaningful features from the query image. For this, we use a DenseNet121~\cite{huang2017densely} model pretrained on ImageNet~\cite{deng2009imagenet}. Importantly, this network is augmented by adding a set of tunable filters between the first layer of the backbone ($7\times7$ convolution layer with stride 2) and the rest of the model. These tunable filters are adjusted by the Object Attention Branch, described below.

\vspace{-1em}
\paragraph{Object attention branch (OAB).}
It has been widely studied that using a pretrained backbone provides better features initialization~\cite{pan2009survey}. For a task related to template matching, this however limits the feature extraction process to be generic and not specialized early on to a particular instance (e.g. it is not necessary to have a high activation on blue objects if we are looking for a red object.). Thus, a specialized branch named ``Object Attention Branch'' (OAB) guides the low-level feature extraction of the backbone network by injecting high-level information pertaining to the object of interest. The output of the OAB can be seen as tunable filters, which are correlated with the feature map of the first layer of the backbone network. The correlation is done within a residual block, similarly to \cite{he2016deep}. The ablation study in sec.~\ref{sec:ablation_network} demonstrates that these tunable filters are instrumental in conferring to a fixed backbone the ability to generalize to objects not seen during training.

The OAB network is a SqueezeNet~\cite{iandola2016squeezenet} pretrained on ImageNet, selected for its relatively small memory footprint and good performance. In order to receive a four-channel input (RGB and binary mask), an extra channel is added to the first convolution layer. The pretrained weights for the first three channels are kept and the weights of the fourth channel are initialized by the Kaiming method~\cite{he2015delving}. During training, a different pose of the target object is sampled at each iteration. For testing, a random pose is sampled once and used on all test images.

\vspace{-1em}
\paragraph{Pose-specific branch (PSB). }
Since an object instance can greatly vary depending on its viewpoint, a ``pose-specific branch'' (PSB) is trained to produce a high-level representation (\emph{embeddings}) of the input object under various poses. This search, based on learned features, is accomplished by depth-wise correlations \emph{and} subtraction with $1\times1$ local templates applied on the backbone output feature map. 
This correlation/subtraction approach is inspired by~\cite{ammirato2018target}, where they have demonstrated an increased detection performance when combining these two operations with $1\times1$ embeddings. Siamese-based object trackers~\cite{bertinetto2016fully,wang2019fast} also use correlations, but with embeddings of higher spatial resolution. We found beneficial to merge these two concepts in our architecture, by using depth-wise correlations (denoted as $\star_d$) in both $1\times1$ and $3\times3$ spatial dimensions. The first one is devoid of spatial information, whereas the second one preserves some of the spatial relationships within a template. We conjecture that this increases sensitivity to orientation, thus providing some cues about the object pose.

This PSB branch has the same structure and weight initialization as the OAB, but is trained with its own specialized weights. The output of that branch are two local template embeddings: at $1\times1$ and $3\times3$ spatial resolution respectively. 
Depth-wise correlations ($1\times1$ and $3\times3$) and subtractions ($1\times1$) are applied between the embeddings generated by this branch and the feature maps extracted from the backbone. All of them are processed by subsequent $3\times3$ convolutions (C1--C3) and are then concatenated.

At test time, the object viewpoint is not known. Therefore, a stack of templates from multiple viewpoints are provided to the pose specific branch. Processing time can be saved at runtime by computing the templates embeddings in an offline phase. Note that the correlation between the local templates and the extracted features is a fast operation and can be easily applied in batch. The backbone network is only processed once per object instance.

\begin{figure*}[t]
\centering
\begin{subfigure}{.3\linewidth}
  \centering
  \includegraphics[width=0.85\linewidth]{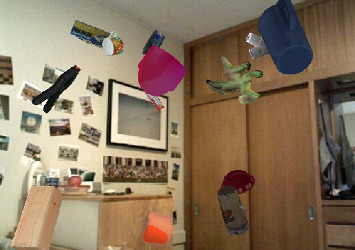}
  \caption{}
\end{subfigure}%
\begin{subfigure}{.3\linewidth}
  \centering
  \includegraphics[width=0.85\linewidth]{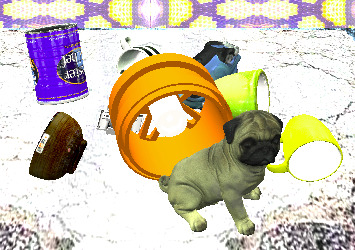}
  \caption{}
\end{subfigure}%
\begin{subfigure}{.3\linewidth}
  \centering
  \includegraphics[width=0.85\linewidth]{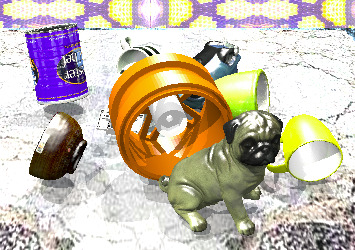}
  \caption{}
\end{subfigure}%
 \vspace{-0.5em}
   \caption{Examples from our domain randomization training set. In (a), objects are randomly placed in front of the camera and rendered using OpenGL with a background sampled from Sun3D dataset~\cite{xiao2013sun3d}. In (b) and (c), a physical simulation is used to drop several objects on a table with randomized parameters (camera position, textures, lighting, materials and anti-aliasing). For each render, 2 variations are saved: one with simple diffuse materials and without shadows (b), and one with more sophisticated specular materials and shadows (c).}
\label{fig_training_data}
\end{figure*}

\subsection{Object detection stage}
\vspace{-0.5em}
The second stage of the network deals with estimating object information from the learned correlation map. The architecture comprises a main task (bounding box prediction) and two auxiliary tasks (segmentation and center prediction). 

\vspace{-1em}
\paragraph{Bounding box prediction.}
The bounding box classification and regression tasks are used to predict the presence and location of the object respectively (as in \cite{lin2017focal}). The classification head predicts the presence/absence of the object for $k$ anchors at every location of the feature map while the regression head predicts a relative shift on the location $(x, y)$ and size $(\text{width}, \text{length})$ with respect to every anchor. In our method, we have $k=24$: 8 scales (30, 60, 90, 120, 150, 180, 210 and 240 pixels) and 3 different ratios (0.5, 1 and 2). Both heads are implemented as 5-layer convolution branches~\cite{lin2017focal}. Inspired from RetinaNet~\cite{lin2017focal}, anchors with an Intersection-over-Union (IoU) of at least 0.5 are considered as positive examples, while those with IoU lower than 0.4 are considered as negatives. The other anchors between 0.4 and 0.5 are not used. At test time, bounding box predictions for all templates are accumulated and predictions with an (IoU) $>$ 0.5 are filtered by Non-Maximum Suppression (NMS). Also, for each bounding box prediction, a depth estimation can be made by multiplying the depth at which the local template was rendered with the size ratio between the local template size (124 pixels) and the prediction size. Predictions that have a predicted depth outside the chosen range of [0.4, 2.0] meters, which is a range that fits to most tabletop settings, are filtered out.

\vspace{-1em}
\paragraph{Segmentation and center prediction.}
The segmentation head predicts a pixel-wise binary mask of the object in the scene image at full resolution. The branch is composed of 5 convolution layers followed by $2\times$ bilinear upsampling layers. Additionally, the center prediction head predicts the location of the object center at the same resolution than the correlation map ($29\times39$) to encourage a strong correlation. The correlation channels are compressed to a single channel heatmap with a $1\times1$ convolution layer.

\vspace{-1em}
\paragraph{Loss Functions.}
The network is trained end-to-end with a main (bounding box detection) and two auxiliary (segmentation and center prediction) tasks. As such, the training loss $\ell_\mathrm{train} = \lambda_1 \ell_\mathrm{seg} + \lambda_2 \ell_\mathrm{center} + \ell_\mathrm{FL} + \ell_\mathrm{reg}$,  
where $\ell_\mathrm{seg}$ is a binary cross-entropy loss for segmentation, $\ell_\mathrm{center}$ is an $L_1$ loss for the prediction of the object center in a heatmap, $\ell_\mathrm{FL}$ is a focal loss~\cite{lin2017focal} associated with the object presence classification and $\ell_\mathrm{reg}$ is a smooth-$L_1$ loss for bounding box regression. The weights $\lambda_1, \lambda_2 $ were empirically set to $20$.



\section{Training data}
\vspace{-0.5em}
In this section, we detail all information related to the input images (query and templates) during training. In particular, we define how the synthetic images are generated and how the dataset is augmented.

\subsection{Domain randomization training images}
\vspace{-0.5em}
We rely on 125 different textured 3D models gathered in majority from the various datasets of the 6D pose estimation benchmark~\cite{hodan2018bop} (excluding Linemod~\cite{hinterstoisser2011multimodal} since it is used for evaluation). Our fully-annotated training dataset is generated with a physic-based simulator similar to~\cite{mitash2017self}, for which objects are randomly dropped on a table in a physical simulation. Every simulation is done in a simple cubic room (four walls, a floor and a ceiling) containing a table placed on the floor in the middle of the room. Inspired from the success of domain randomization~\cite{tobin2017domain,tremblay2018training}, we added randomness to the simulation parameters in order to reduce the domain gap between synthetic and real images. 
The following parameters are randomized: the texture of the environment (walls, floor and table), lighting (placement, type, intensity and color), object materials (diffuse and specular reflection coefficients) and anti-aliasing (type and various parameters).

\vspace{-1em}
\paragraph{Renders.} Our physics-based domain randomization dataset is composed of 10,000 images. To generate these images, we ran 250 different simulations with different sets of objects (between 4 and 13 objects in each simulation). In 50\% of the simulations, objects were automatically repositioned to rest on their bottom/main surface to replicate a bias found in many tabletop datasets. For each simulation, 20 camera positions were randomly sampled on half-spheres of radius ranging from 0.8 to 1.4 meters, all pointing towards the table center with random offsets of $\pm 15$ degrees for each rotation axis. For each sampled camera position, two image variations were rendered: one with realistic parameters (containing reflections and shadows) as shown in fig.~\ref{fig_training_data}-(c) and the other without, as shown in fig.~\ref{fig_training_data}-(b). Tremblay et al.~\cite{tremblay2018deep} showed that using different kinds of synthetic images reduced the performance gap between synthetic and real images. Accordingly, we have generated an additional set of 10,000 simpler renders using OpenGL. For this, we rendered objects in random poses on top of real indoor backgrounds sampled from the Sun3D dataset~\cite{xiao2013sun3d} (fig.~\ref{fig_training_data}-(a)).

\paragraph{Labels.} After the simulations, we kept the 6 degree of freedom pose of each object as the ground truth. We used the pose together with the 3D model to generate a visibility mask for the segmentation task, and projected the center of the 3D model in the image plane to generate the center heatmap. The ground-truth heatmap is a 2D Gaussian with an amplitude of 1 and a variance of 5 at the projected center of the object at an image resolution equivalent to the output of the network.

\subsection{Templates}
\vspace{-0.5em}
The following section describes the template generation procedure for training. We also remind the different procedure used at test time, as described in~sec.~\ref{sec:correlation_stage}.

For each training iteration, one of the objects from the query image is selected as the target object and all the others are considered as background. All templates are rendered with a resolution of $124\times124$ pixels. To render consistent templates from multiple objects of various size, we adjust the distance of the object so that its largest length on the image plane falls in the range of 100 to 115 pixels. The borders are then padded to reach the size of $124\times124$.

\paragraph{Global template (OAB):}
In an offline phase, 240 templates are generated for each 3D model by sampling 40 viewpoints on an icosahedron with 6 in-plane rotations per viewpoint. During training, one of the 240 templates is sampled randomly for each iteration. At test time, a single one is randomly selected for all experiments.

\paragraph{Local template (PSB):}
We apply perturbations on the orientation of the template image by sampling a random rotation axis and rotation magnitude, and adding that perturbation to the ground truth viewpoint before rendering the local template. The impact of using different rotation magnitude is quantified in table~\ref{tab:ablation_rotation}, with best performance obtained with random rotations perturbation in the range of 20--30$^\circ$ to the ground truth viewpoint. At test time, a stack of 160 templates rendered from 16 viewpoints is used.

\subsection{Data augmentation}
\vspace{-0.5em}
Online data augmentation is applied to synthetic images during training. We use the segmentation mask of the object in the query image to randomly change the hue, saturation and brightness of the object and its template. We also apply augmentations on the whole query image, such as: brightness shifts, Gaussian blur and noise, horizontal and vertical flips, random translations and scale.
To minimize the risk of overfitting to object color, a random hue is applied to the whole image and the template 50\% of the time.
Finally, we apply motion blur to the image 20\% of the time by convolving a line kernel to the image, as in~\cite{Dwibedi_2017_ICCV}.

\section{Experiments}
\label{sec_experiments}

\vspace{-0.5em}
In this section, we provide details on the training procedure and on the dataset and metrics used to evaluate our approach. We also describe the various ablation studies that validate our design choices. Finally, we present an extensive evaluation against the state-of-the-art methods.

\subsection{Training details} 
\vspace{-0.5em}
Our complete network is trained for 50 epochs with AMSGrad~\cite{reddi2019convergence}.
We used a learning rate of $10^{-4}$ with steps of 0.1 at epochs 20 and 40, a weight decay of $10^{-6}$ and mini batches of size 6. We used 1k renders as a validation set and used the remaining 19k of the generated dataset (OpenGL and physics-based) for training. Each epoch, the network was trained for 1,300 iterations and images are sampled with a ratio of 80/20 respectively from the physics-based and OpenGL renders. Once the training was complete, the network with the smallest validation loss (computed at the end of each epoch) was kept for testing. 

\subsection{Datasets and metrics}
\label{sec:datasets_metrics}
\vspace{-0.5em}
We evaluate on the well-known Linemod~\cite{hinterstoisser2012model} and Occluded Linemod~\cite{brachmann2014learning} datasets. Linemod consists of 15 sequences of real objects containing heavy clutter where the annotations of a single object are available per sequence. Occluded Linemod is a subset of Linemod, where annotations for 8 objects have been added by~\cite{brachmann2014learning}. Keeping in line with previous work, we only keep the prediction with the highest score for each object and use the standard metrics listed below. We use a subset containing 25\% of the Linemod dataset for the ablation studies.

\vspace{-1em}
\paragraph{Linemod.} The standard metric for this dataset is the ``2D bounding box''~\cite{brachmann2014learning}, which calculates the ratio of images for which the predicted bounding box has an intersection-over-union (IoU) with the ground truth higher than 0.5. 

\paragraph{Occluded Linemod.} The standard mean average precision (mAP) is used to evaluate the performance of multi-object detection. To allow for direct comparison, we regroup the predictions made for different objects and apply NMS on predictions with an IoU $>$ 0.5. We use the same methodology as in~\cite{brachmann2014learning}: the methods are evaluated on 13 of the 15 objects of the Linemod dataset (the ``bowl'' and ``cup'' objects are left out). Of the remaining 13 objects, 4 are never found in the images, yet those are still detected and kept in the evaluation (as an attempt to evaluate the robustness to missing objects). The mAP is therefore computed by using all the predictions on the 9 other objects left.

\subsection{Ablation studies}
\label{sec:ablation_network}

\paragraph{Network architecture.}
We evaluate the importance of different architecture modules (presented in sec.~\ref{sec:architecture}). For each test, a specific module is removed and its performance is compared to the full architecture.
Tab.~\ref{tab:ablation_architecture} shows that removing the ``Object Attention Branch'' resulted in the largest performance drop (almost 20\%). Also, removing the higher-resolution $3\times3$ embeddings and auxiliary tasks reduced performances by approximately 5\% and 8\% respectively. 

\begin{table}[!t]
\centering
\small
\begin{tabular}{lc}
\toprule
Network $\quad$& $\Delta$ performance (\%) \\
\midrule
w/o tunable filters (OAB) $\quad$& -19.76 \\
w/o auxiliary tasks $\quad$& $\;  $  -7.73  \\
w/o $3\times3$ correlation (PSB) $\quad$& $\;  $ -5.37  \\
\bottomrule
\end{tabular}
\vspace{-0.5em}
\caption{Network architecture ablation study. Removing tunable filters resulted in the most notable performance drop.}
\label{tab:ablation_architecture}
\vspace{-0.1in}
\end{table}

\vspace{-1em}
\paragraph{Importance of local template perturbation during training.}
A perfect match between the template pose and the target object pose in the scene is unlikely. As such, the training procedure must take this into account by adding orientation perturbations to local templates at train time. Here, we investigated what is the desirable magnitude of such perturbations.
In tab.~\ref{tab:ablation_rotation}, a random rotation of $0^\circ$ represents local templates selected with the same orientation as the object in the scene. Perturbations are then added by randomly sampling a rotation axis (in spherical coordinates) and a magnitude. A network was retrained for each amount of perturbation. 
Tab.~\ref{tab:ablation_rotation} illustrates that perturbations of 20--30$^\circ$ seems to be optimal. Networks trained with too small perturbations may not be able to detect objects under all their possible configurations, resulting in small performances drop of less than 5\%, and those trained with too big perturbations are more prone to false detections (the network is trained to allow for bigger differences in appearance and shape between the template and scene object), resulting in a bigger drop of 16\% for rotations of 180$^\circ$.

\begin{table}
\small
\centering
\begin{tabular}{cc}
\toprule
Random rotations & $\Delta$ performance (\%) \\
\midrule
0$^\circ$ & $\;$ -4.33 \\
$\pm$ 10$^\circ$ & $\;$ -3.12 \\
$\pm$ 20$^\circ$  & $\;$  0 \\
$\pm$ 30$^\circ$ & $\;$ -0.42  \\
$\pm$ 40$^\circ$  & $\;$ -5.18  \\
$\pm$ 180$^\circ$ & -16.07  \\
\bottomrule
\end{tabular}
\vspace{-0.5em}
\caption{Impact of perturbing the pose of local templates (instead of using the the ground truth pose) during training.}
\label{tab:ablation_rotation}
\end{table}

\begin{table}
\small
\centering
\begin{tabular}{ccc}
\toprule
\# of templates & $\Delta$ performance (\%) & runtime (ms) \\
\midrule
80 & -2.80 & 230 \\
160 & 0.00 & 430 \\
320 & +0.03 & 870 \\
\midrule
1 (oracle) & +16.75  & 60\\
\bottomrule
\end{tabular}
\vspace{-0.5em}
\caption{Bounding box detection performance and runtime for various numbers of local templates at test time. The oracle sets an upper bound of performance by providing a single template with the ground truth object pose.}
\label{tab:ablation_templates}
\end{table}

\vspace{-0.5em}
\paragraph{Number of local templates.} The impact of providing various numbers of local templates to the network at test time is evaluated, both in terms of accuracy and speed, in tab.~\ref{tab:ablation_templates}. Timings are reported on a Nvidia GeForce GTX 1080Ti. To generate a varying number of templates, we first selected 16 pre-defined viewpoints spanning a half-sphere over the object. Each template subsequently underwent 5 (80 templates), 10 (160 templates) and 20 (320 templates) in-plane rotations. Tab.~\ref{tab:ablation_templates} compares performance with that obtained with an oracle who used a template with the ground truth pose. Performance ceases to improve beyond 160 templates.

\paragraph{Global template selection.} In tab.~\ref{tab:ablation_global_template}, we show that the object pose of the global template does not impact significantly the detection performance. For the first test, we report the average performance of 5 different evaluations in which a random global template was selected. The performance slightly improved compared to the random template used in all other tests, suggesting that the template selection in every other test was suboptimal. However, it also shows that the templates were not cherry-picked for optimal performance on the test datasets. Secondly, we show that using a template of the good object is primordial. Using empty templates (all 0's) or providing templates from another object results in a dramatic performance drop of more than 30\%, thus hinting about the discriminative power of the OAB.

\begin{table}[!t]
\small
\centering
\begin{tabular}{lc}
\toprule
Global template selection & $\Delta$ performance (\%) \\
\midrule
Random Pose & +1.21\\
Empty & -32.47 \\
Wrong Object & -38.15\\
\bottomrule
\end{tabular}
\vspace{-0.5em}
\caption{Robustness towards different selection of global templates at test time.}
\label{tab:ablation_global_template}
\vspace{-0.1in}
\end{table}

\begin{figure}[!t]
\begin{center}
\begin{tabular}{cc}
\includegraphics[width=0.42\linewidth]{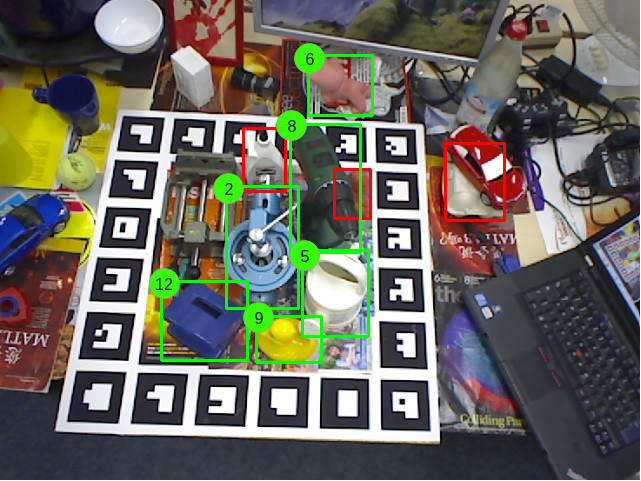} &
\includegraphics[width=0.42\linewidth]{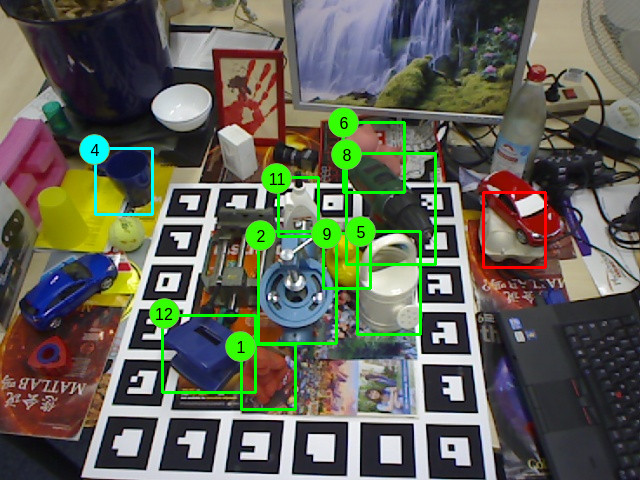} \\
\includegraphics[width=0.42\linewidth]{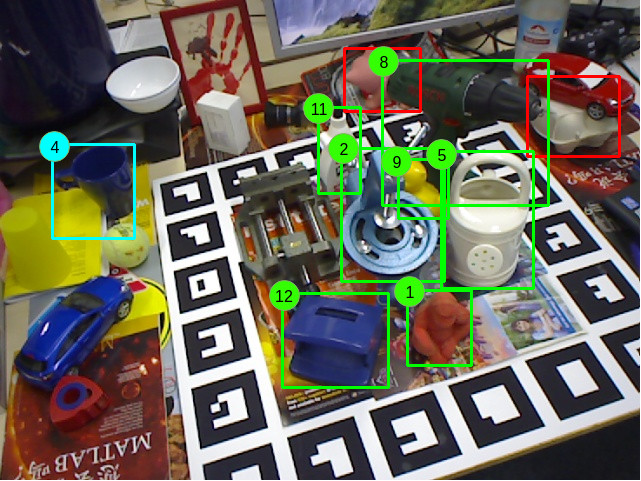} &
\includegraphics[width=0.42\linewidth]{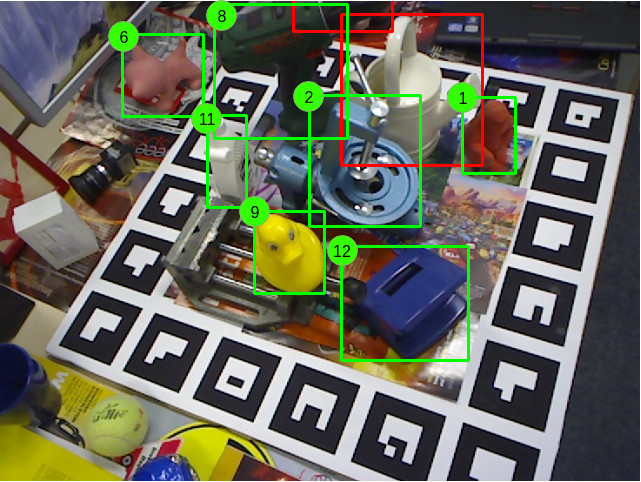} \\
\end{tabular}
\includegraphics[width=0.84\linewidth]{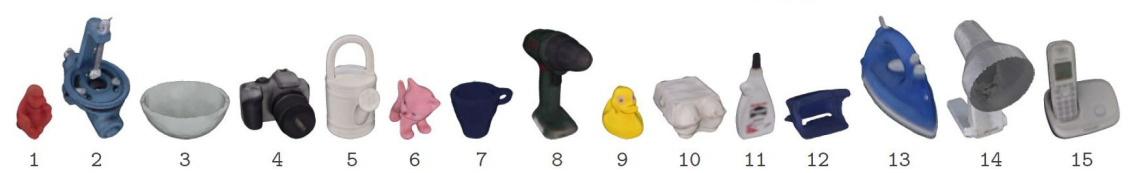}
\end{center}
   \vspace{-0.2in}
   \caption{Qualitative results on the Occluded Linemod dataset~\cite{brachmann2014learning}, showing good (green), false (blue) and missed (red) detections. For reference, the 15 objects are shown in the bottom row (image from~\cite{hodan2018bop}). Here, all objects (except objects 3 and 7) are searched in each image.}
\label{fig:results_occluded}
\end{figure}

\paragraph{Number of objects in the training set.}
The network was retrained on subsets of objects of the synthetic dataset. The remaining objects were considered as background clutter. Tab.~\ref{tab:number_of_objects} shows the performance w.r.t the quantity of objects used during training. While using few objects still performs reasonably well, 
more objects does improve generalization.

\begin{table}[!t]
\small
\centering
\begin{tabular}{cc}
\toprule
\# of objects & $\Delta$ performance (\%) \\
\midrule
15 & -31.58 \\
30 & -15.97 \\
63 & -10.42 \\
90 & -3.58 \\
125 & 0 \\
\bottomrule
\end{tabular}
\vspace{-0.5em}
\caption{Impact of the number of objects used in training.}
\label{tab:number_of_objects}
\vspace{-1.5em}
\end{table}

\paragraph{Similar objects in the training set.}
While no single object were present in both training and test sets, it is possible that the training set contained objects that shared similarities to objects in the test set. To evaluate the potential impact this might have, we removed all cups from our training set (13 were found), trained a network on the resulting set, and evaluated its performance on the test set. Doing so reduced the overall score by less than 1\%, but the average performance solely on cups slightly improved (not statistically significant). This experiment demonstrates that the network does not overfit on a particular class of instances.

\begin{table}[!t]
\centering
\footnotesize
\begin{tabular}{lcccc}
\toprule
 \multirow{2}{*}{Methods} & Known & Real & Linemod & OL \\
  & objects & images & (2D BBox) & (mAP) \\
\midrule
Brachmann et al.~\cite{brachmann2016uncertainty} 	& Yes 	& Yes		& 97.50		& 51.00 \\
SSD-6D \cite{kehl2017ssd}							& Yes  	& No		& 99.40		& 38.00 \\
DPOD~\cite{zakharov2019dpod}						& Yes  	& No		& N/A		& 48.00 \\
Hodan et al.~\cite{hodan2019photorealistic}			& Yes  	& No		& N/A		& 55.90* \\
\midrule
Tjaden et al.~\cite{tjaden2017real} 				& No	& Yes 		& 78.50		& N/A \\
LINE-2D~\cite{hinterstoisser2011multimodal}			& No  	& No		& 86.50 	    & 21.0 \\
TDID corrs.~\cite{ammirato2018target}         & No  	& No		& 54.37 	& 34.13 \\
SiamMask corrs.~\cite{wang2019fast}           & No  	& No		& 68.23 	& 41.47 \\
Ours 												& No 	& No		& 77.92 	    & 50.71 \\
\bottomrule
\end{tabular}
\vspace{0.25em}
\caption{Quantitative comparison to the state of the art, with 2D bounding box metric on Linemod and mAP on Occluded Linemod (OL). The 2D bounding box metric calculates the recall for the 2D bounding boxes with the highest prediction score. For both metrics, predictions are considered good if the IoU of the prediction and the ground truth is at least 0.5 (0.75 for Hodan et al.~\cite{hodan2019photorealistic}). The methods are separated first according to their prior knowledge of test objects and then if real images similar to the test set are used to optimize the performance. Our approach is the most robust of all methods that were not trained for the test objects, having a good score on Linemod and the best score on Occluded Linemod.}
\label{tab:main_results}
\vspace{-1.5em}
\end{table}

\subsection{Comparative evaluation to the state of the art}
\vspace{-0.5em}
We report an evaluation on Linemod and Occluded Linemod (OL) in tab.~\ref{tab:main_results} and compare with other state-of-the-art RGB-only methods. Competing methods are divided into 2 main groups: those who do know the test objects at train time (``known objects''), and those who do not. Approaches such as \cite{brachmann2016uncertainty,kehl2017ssd,zakharov2019dpod,hodan2019photorealistic} are all learning-based methods that were specifically trained on the objects. On the other hand, \cite{tjaden2017real,hinterstoisser2011multimodal} and~\cite{ammirato2018target,wang2019fast} are respectively template matching and learning-based methods that do not include a specific training step targeted towards specific object instances. It is worth noting that even though \cite{tjaden2017real} is classified as not needing known objects at training time, it still requires an initialization phase using real images (to build a dictionary of histogram features). As in \cite{brachmann2016uncertainty}, they thus use parts of the Linemod dataset as a training set that covers most of the object viewpoints. These methods have therefore an unfair advantage compared to our approach and Line-2D, since they leverage domain-specific information (lighting, camera, background) of the evaluation dataset.

Our method is evaluated without prior knowledge of the Linemod objects. It can be directly compared with Line-2D~\cite{hinterstoisser2011multimodal} which also uses templates as input.
On the standard Linemod dataset, Line-2D outperforms our method by 8.5\% on the ``2D bounding box'' metric.
The better results of Line-2D on Linemod can be explained in part by an additional and naive post-processing color-based check that rejects false positives~\cite{brachmann2016uncertainty} while we report the performance of our approach without any post-processing. We note that this naive approach fails if minor occlusions occurs.
In contrast, our method outperforms Line-2D by almost 30\% in mAP on the more difficult Occluded Linemod. 
Our approach also provides competitive performance that is on par or close to all other methods that test on known objects and/or have access to real images.
Fig.~\ref{fig:results_occluded} shows qualitative results on Occluded Linemod.
We also compare our approach with TDID~\cite{ammirato2018target} and SiamMask~\cite{wang2019fast}. We replaced their original backbones by the same architecture (DenseNet) we are using. As specified by those methods, a siamese backbone replaced our 2 branches approach (OAB and PSB). TDID uses a $1\times1$ embedding whereas a $3\times3$ embedding is used for SiamMask. All implementations were trained on the same task following the same procedure than our approach and their scores are reported in tab.~\ref{tab:main_results}. Overall, our proposed approach significantly outperforms these two baselines.
\section{Discussion}
\vspace{-0.5em}
We have proposed a method for detecting specific object instances in an image that does not require knowledge of the object at training time. At test time, the proposed network takes multiple viewpoints of the object as input, and predicts its location from a single RGB image. Our experiments show that while the network has not been trained with any of the test objects, it is significantly more robust to occlusion than previous template-based methods (30\% improvement in mAP over Line-2D~\cite{hinterstoisser2011multimodal}). It is also highly competitive with networks that are specifically trained on the object instances.

\paragraph{Limitations.} False positives arise from clutter with similar color/shape as the object, as shown in fig.~\ref{fig:results_occluded}. We hypothesize that our late correlation at small spatial resolution (templates of $3\times3$ and $1\times1$) prevents the network from leveraging detailed spatial information related to the object shape. Another limitation is that the method requires 0.43s to detect a single object instance in an image (c.f. tab.~\ref{tab:ablation_templates}), scaling linearly with the number of objects. The main reason for this is the object attention branch (OAB), which makes the backbone features instance-specific via tunable filters, which needs to be recomputed for each object.
Also, while capturing a 3D model has become increasingly simpler (it takes less than 5 minutes with commodity hardware~\cite{hinterstoisser2019annotation}), this may not always be practical. While our experiments rely on such 3D models to allow for quantitative evaluation on standard datasets for which only the 3D model is available, obtaining multiple viewpoints of an object could also be done simply by photographing it against a uniform background. 

\paragraph{Future directions.} By providing a generic and robust 2D instance detection framework, this work opens the way for new methods that can extract additional information about the object, such as its full 6-DOF pose. We envision a potential cascaded approach, which could first detect unseen objects, and subsequently regress the object pose from a high-resolution version of the detection window.

\section*{Acknowledgements} 

{\small This work was supported by the REPARTI Strategic Network and the NSERC/Creaform Industrial Research Chair on 3D Scanning: CREATION 3D and Nvidia.}

{\small
\bibliographystyle{ieee_fullname}
\bibliography{egbib}
}

\newpage

\appendix


\section{Per object performances }
\begin{figure}[h]
\centering
\includegraphics[width=\linewidth]{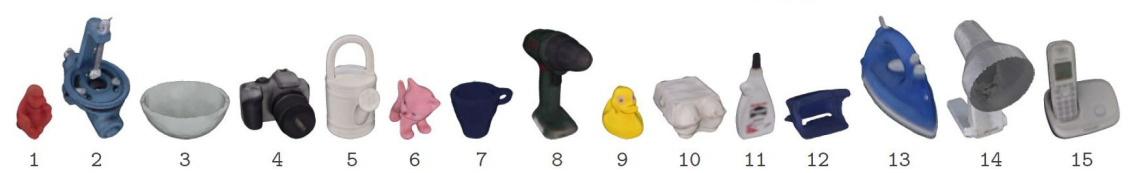}
\caption{All 15 objects in the Linemod dataset (taken from~\cite{hodan2018bop}).}
\label{fig_linemod_objects_supp}
\end{figure}

In the main paper, we reported the performance of our approach on Linemod~\cite{hinterstoisser2012model} and Occluded Linemod~\cite{brachmann2014learning} datasets. We extend the reported results by showing the performance of our approach on each object of both datasets in tab.~\ref{tab_linemod_per_obj} and tab.~\ref{tab_occluded_linemod_per_obj}. Linemod objects, along with their corresponding indices, can be viewed in fig.~\ref{fig_linemod_objects_supp}.

\begin{table}[ht]
\centering
\begin{tabular}{cc}
\toprule
Object ID & 2D BBox metric (\%) \\ 
\midrule
1                             & 89.16                                   \\ 
2                             & 71.50                                   \\ 
3                             & 94.00                                   \\ 
4                             & 46.88                                   \\ 
5                             & 92.47                                   \\ 
6                             & 80.75                                   \\ 
7                             & 82.74                                   \\ 
8                             & 77.19                                   \\ 
9                             & 63.31                                   \\ 
10                            & 96.89                                   \\ 
11                            & 89.51                                   \\ 
12                            & 67.83                                   \\ 
13                            & 87.67                                   \\ 
14                            & 86.39                                   \\ 
15                            & 42.48                                   \\ 
Mean                          & 77.92                                    \\ 
\bottomrule 
\end{tabular}
\vspace{0.5em}
\caption{Performances on the 2D bounding box metric for each object of the Linemod dataset.}
\label{tab_linemod_per_obj}
\end{table}

\begin{table}[ht]
\centering
\begin{tabular}{cc}
\toprule
Object ID & Mean Average Precision (mAP) \\
\midrule
1                             & 36.58                                      \\ 
2                             & 55.92                                      \\ 
5                             & 73.49                                      \\ 
6                             & 29.18                                      \\ 
8                             & 55.20                                      \\ 
9                             & 77.48                                      \\ 
10                            & 52.79                                      \\ 
11                            & 16.26                                      \\ 
12                            & 59.52                                      \\ 
mAP & 50.71                                      \\ 
\bottomrule
\end{tabular}
\vspace{0.5em}
\caption{Average precision for each object evaluated on the Occluded Linemod dataset.}
\label{tab_occluded_linemod_per_obj}
\end{table}

\pagebreak
\section{Domain randomization training images}
\begin{figure*}[h]
\begin{center}
\begin{tabular}{ccc}
\includegraphics[width=0.3\linewidth]{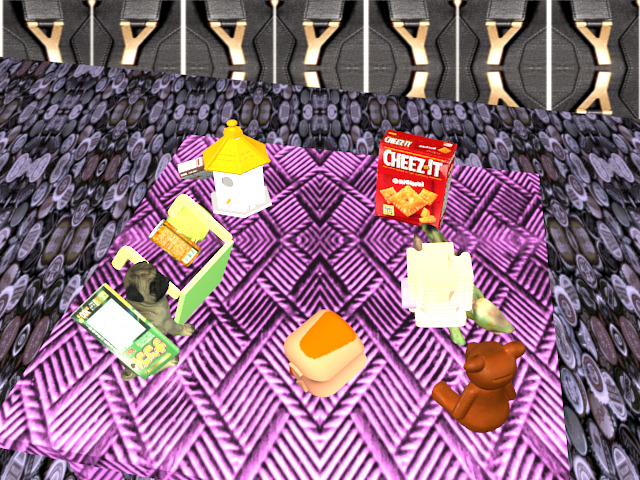} &
\includegraphics[width=0.3\linewidth]{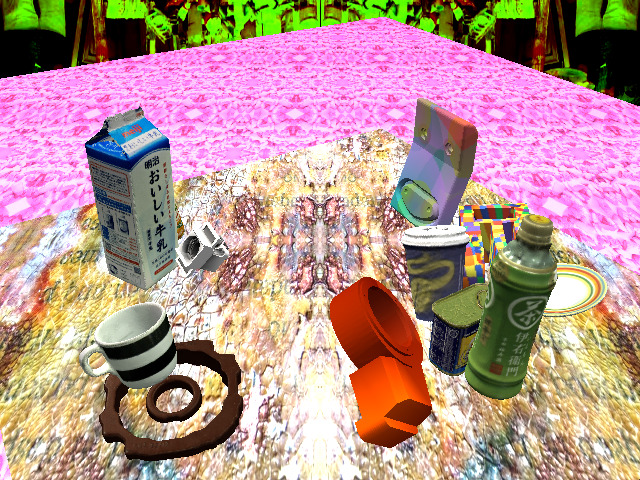} &
\includegraphics[width=0.3\linewidth]{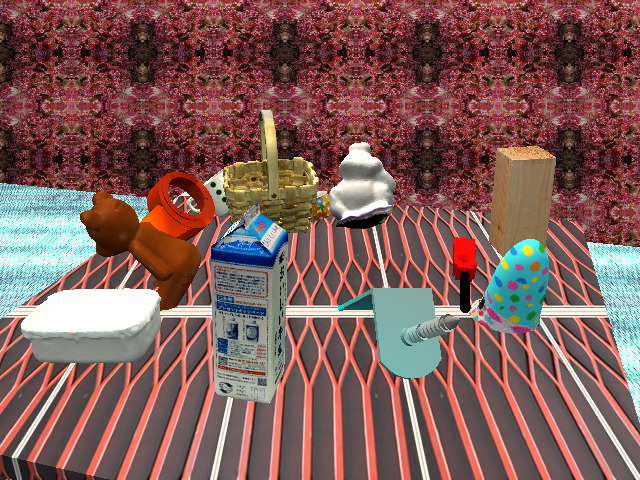} \\
\includegraphics[width=0.3\linewidth]{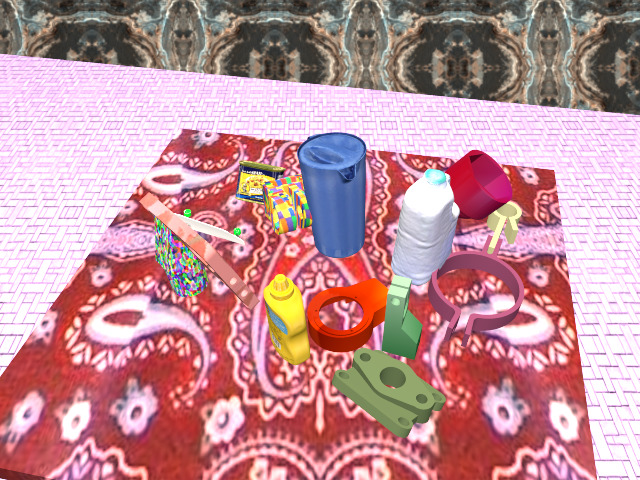} &
\includegraphics[width=0.3\linewidth]{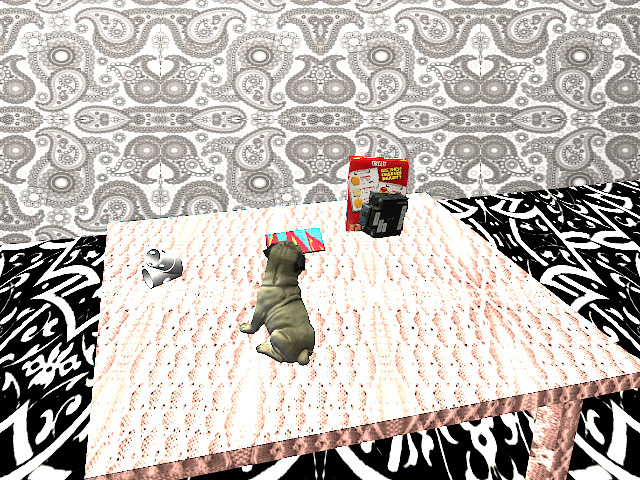} &
\includegraphics[width=0.3\linewidth]{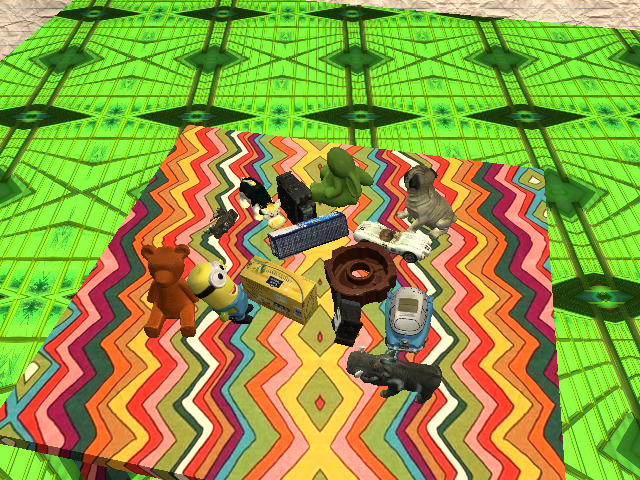} \\
\includegraphics[width=0.3\linewidth]{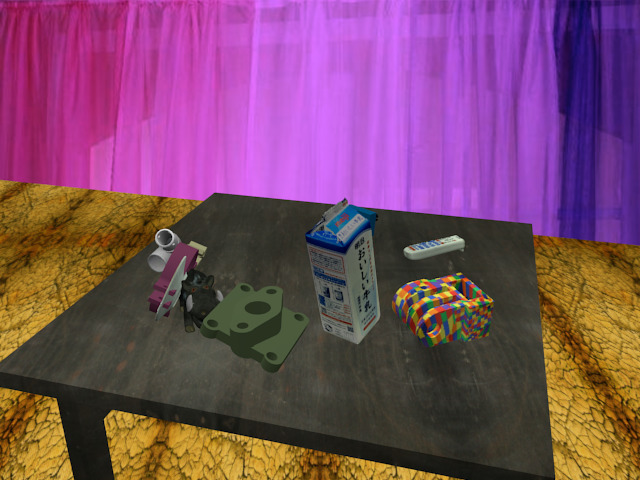} &
\includegraphics[width=0.3\linewidth]{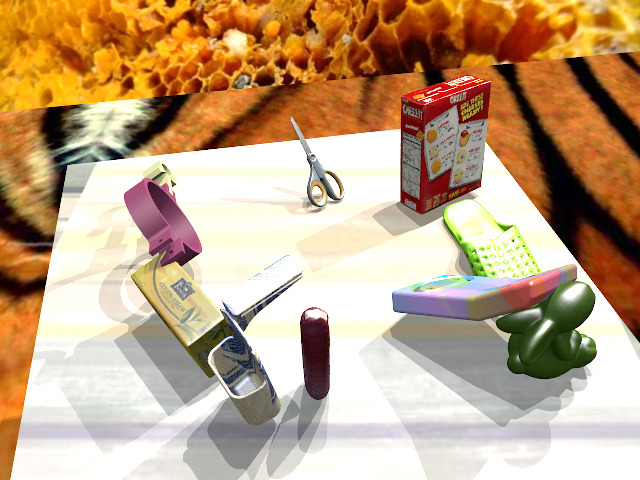} &
\includegraphics[width=0.3\linewidth]{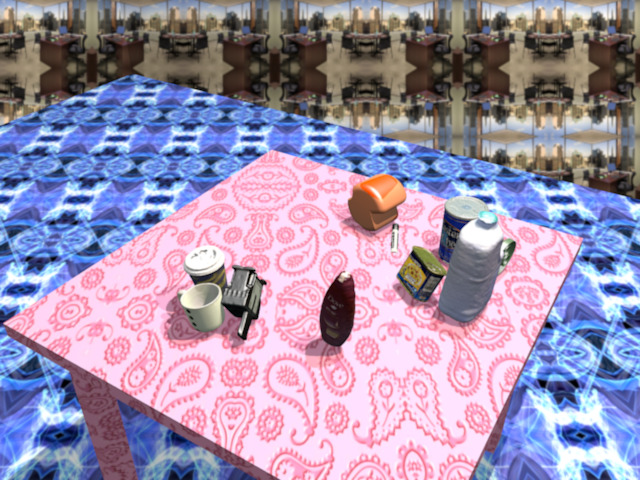} \\
\includegraphics[width=0.3\linewidth]{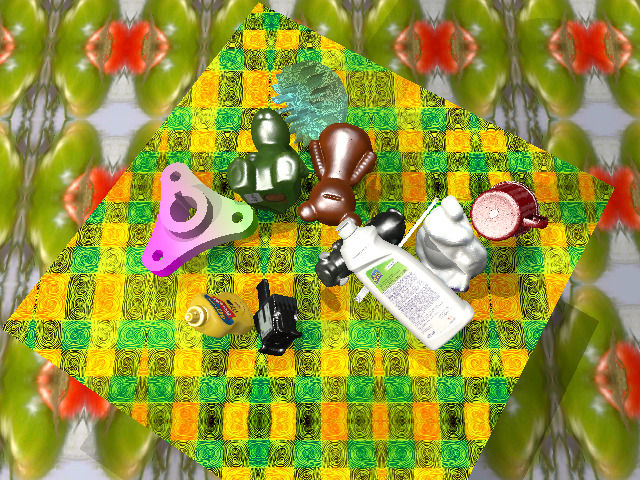} &
\includegraphics[width=0.3\linewidth]{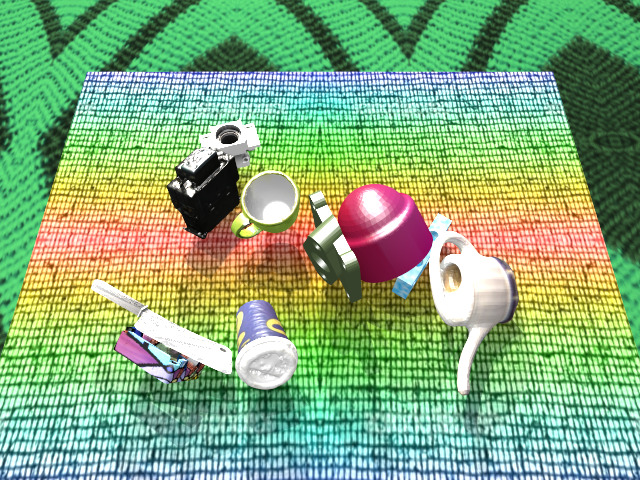} &
\includegraphics[width=0.3\linewidth]{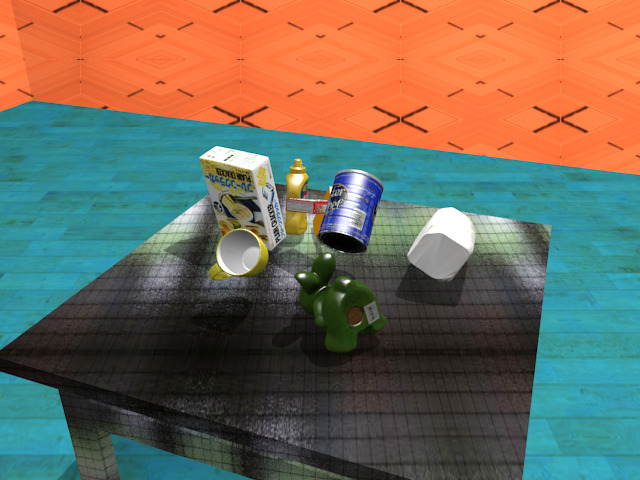} \\
\end{tabular}

\end{center}
   \caption{More domain randomization images generated with our simulator}
\label{fig_dom_rand}
\end{figure*}

Additional examples of domain randomization images generated with our simulator are shown in fig.~\ref{fig_dom_rand}. 

\section{Qualitative results on Linemod dataset}
\begin{figure*}[h]
\begin{center}
\begin{tabular}{ccc}
\includegraphics[width=0.3\linewidth]{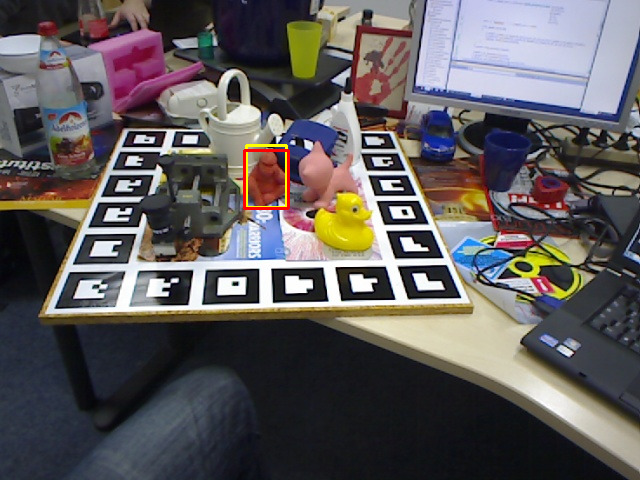} &
\includegraphics[width=0.3\linewidth]{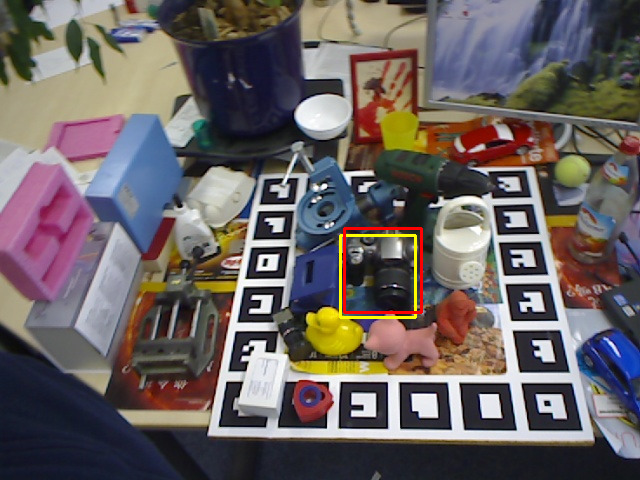} &
\includegraphics[width=0.3\linewidth]{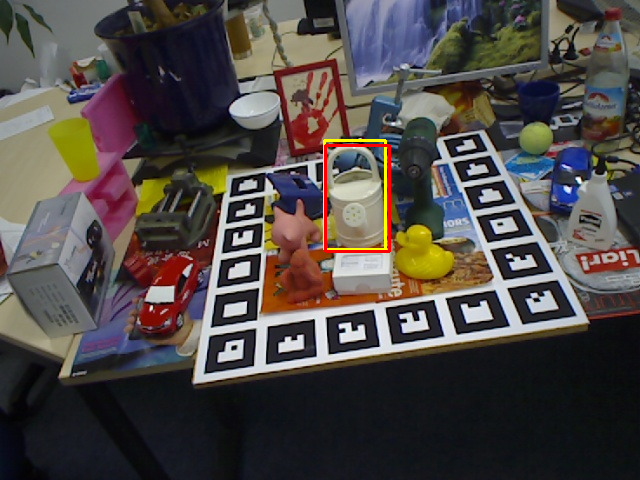} \\
\includegraphics[width=0.3\linewidth]{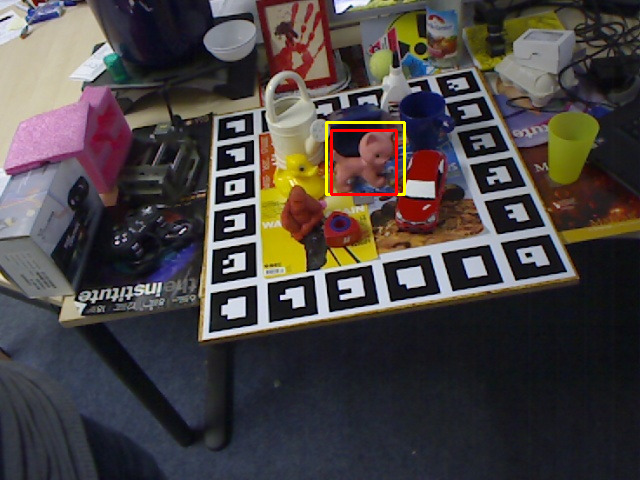} &
\includegraphics[width=0.3\linewidth]{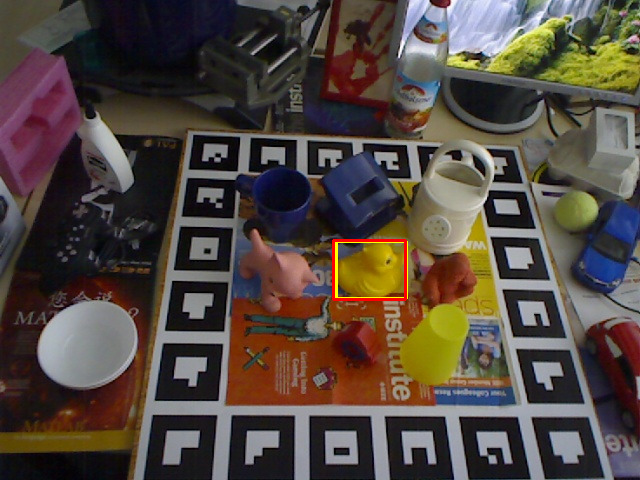} &
\includegraphics[width=0.3\linewidth]{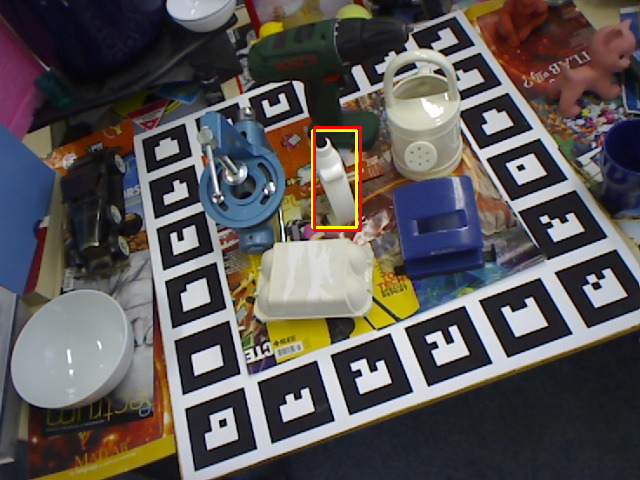} \\
\includegraphics[width=0.3\linewidth]{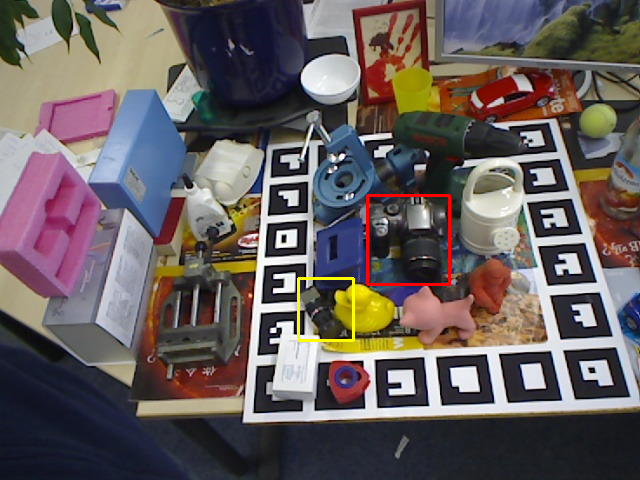} &
\includegraphics[width=0.3\linewidth]{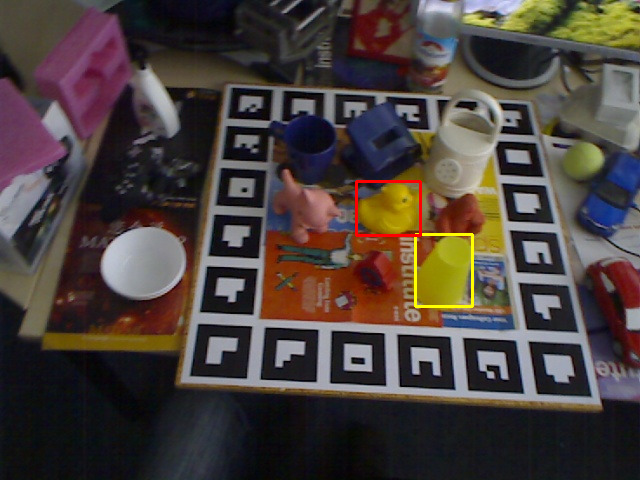} &
\includegraphics[width=0.3\linewidth]{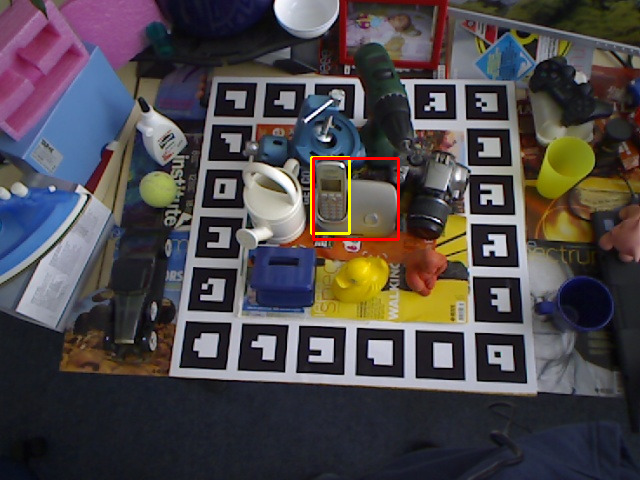} \\

\end{tabular}

\end{center}
   \caption{Qualitative results on Linemod dataset~\cite{hinterstoisser2012model} with predictions (yellow) and ground-truths (red). The first two rows show good predictions while the last row shows examples of bad predictions.}
\label{fig_results_linemod}
\end{figure*}

We show examples of good and bad predictions on Linemod dataset in fig.~\ref{fig_results_linemod}. 

\section{Qualitative results on Occluded Linemod}
\begin{figure*}[thpb]
\begin{center}
\begin{tabular}{cc}
\includegraphics[width=0.45\linewidth]{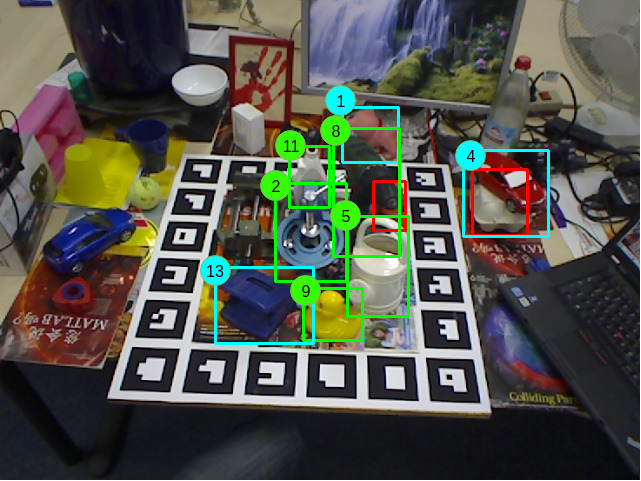} &
\includegraphics[width=0.45\linewidth]{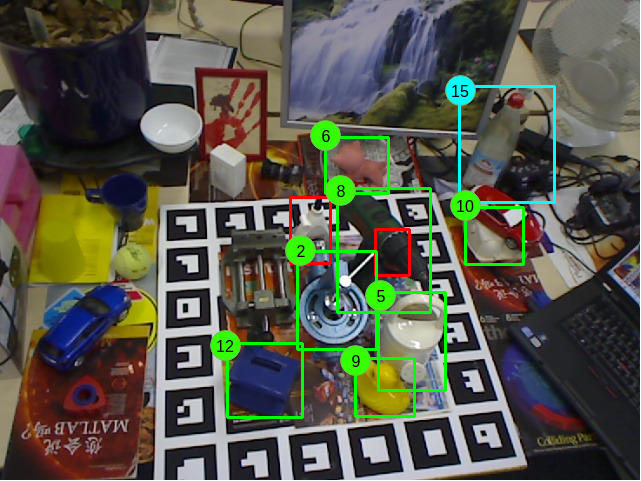} \\
\includegraphics[width=0.45\linewidth]{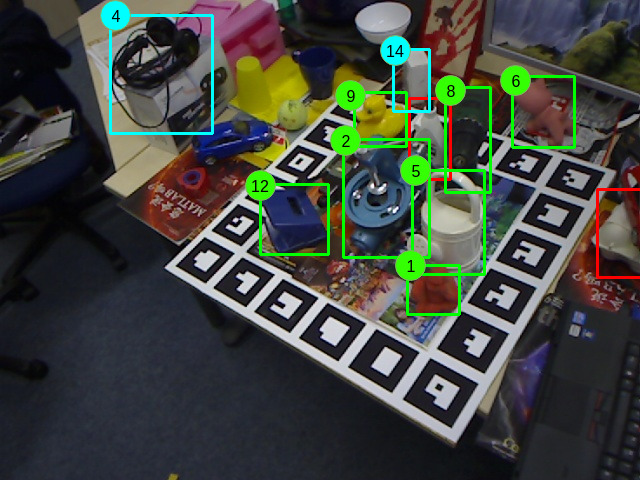} &
\includegraphics[width=0.45\linewidth]{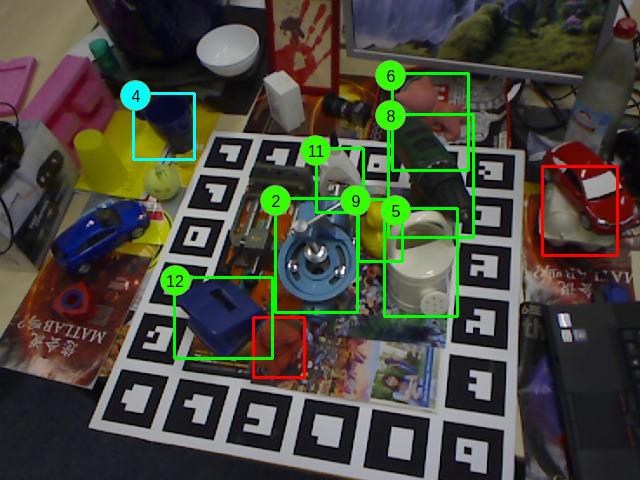} \\
\includegraphics[width=0.45\linewidth]{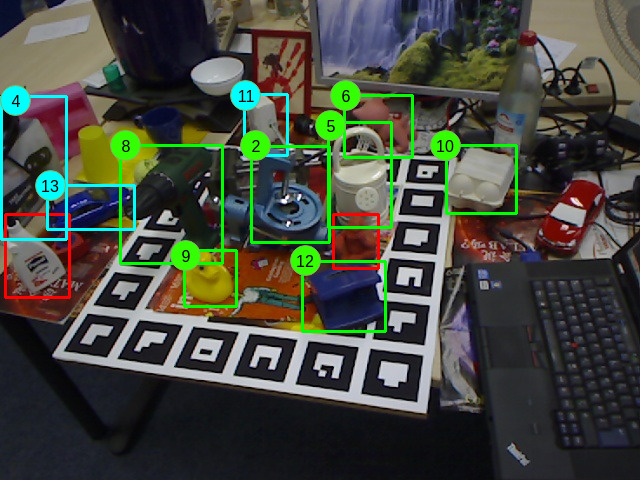} &
\includegraphics[width=0.45\linewidth]{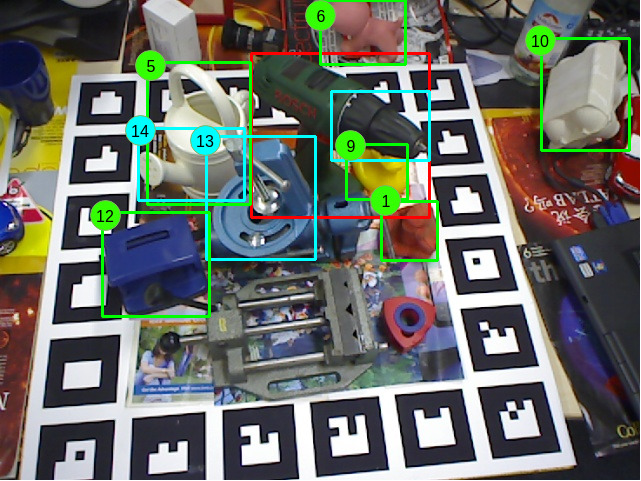} \\
\end{tabular}
\end{center}
   \caption{More qualitative results on the Occluded Linemod dataset~\cite{brachmann2014learning}, showing good (green), false (blue) and missed (red) detections.}
\label{fig_results_occluded_supp}
\end{figure*}

We show additional qualitative results on Occluded Linemod in fig.~\ref{fig_results_occluded_supp} to expand results shown in the paper.

\section{Architecture details}
\begin{figure*}[thpb]
\begin{center}
\begin{tabular}{c}
\includegraphics[width=0.75\linewidth]{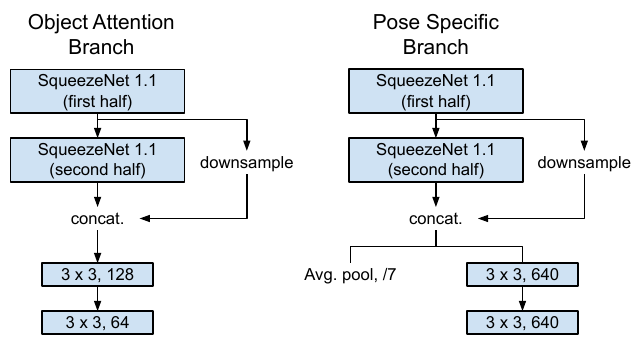} \\
\toprule
\includegraphics[width=0.75\linewidth]{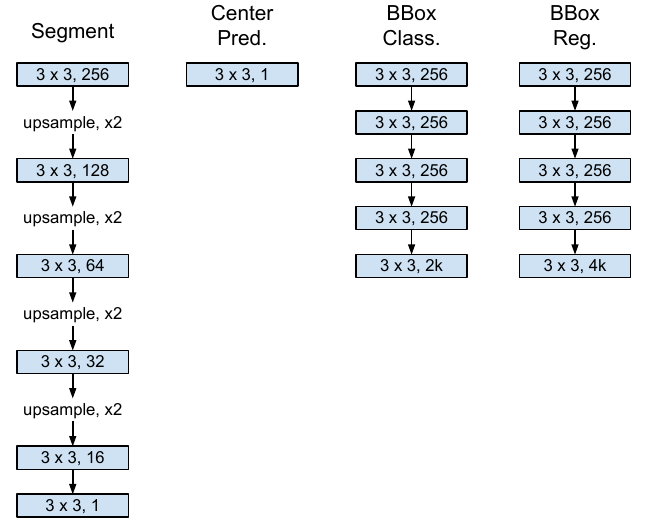} \\
\end{tabular}
\end{center}
   \caption{Detailed networks}
\label{fig_detailed_arch}
\end{figure*}

Each subnetwork is shown in details in fig~\ref{fig_detailed_arch}.

\end{document}